\documentclass{IOS-Book-Article}

\usepackage{mathptmx}
\usepackage{graphicx} 
\usepackage{makecell} 

\usepackage[T1]{fontenc}
\usepackage{xcolor}
\usepackage{colortbl}
\usepackage{tikz}
\usepackage{booktabs}
\usepackage{tabularx}   
\usepackage{makecell}
\usepackage{graphicx}

\usepackage{soul}\setuldepth{article}
\def\hb{\hbox to 11.5 cm{}}

\newcommand{\yes}{Y}
\newcommand{\partialm}{P}
\newcommand{\no}{N}
\begin{document}

\pagestyle{headings}
\def\thepage{}
\begin{frontmatter}              

\title{OntoExtend: A Framework for Requirement-driven and Scalable Ontology Extension with LLMs}

\markboth{}{April 2026\hb}

\author[B,D]{\fnms{Anna Sofia} \snm{Lippolis}\textsuperscript{*}}
\author[A]{\fnms{Mohammad Javad} \snm{Saeedizade}\textsuperscript{*}}
\author[C]{\fnms{Stefan} \snm{Schmid}}
\author[C]{\fnms{Simon} \snm{Blattner}}
\author[A]{\fnms{Robin} \snm{Keskisärkkä}}
\author[B]{\fnms{Aldo} \snm{Gangemi}}
\author[A]{\fnms{Eva} \snm{Blomqvist}}
\author[D]{\fnms{Andrea Giovanni} \snm{Nuzzolese}}

\runningauthor{Lippolis et al.}

\address[A]{Linköping University, Sweden}
\address[B]{University of Bologna, Italy}
\address[C]{Bosch, Germany}
\address[D]{ISTC-CNR, Italy}

\noindent\textsuperscript{*}These authors contributed equally to this work.

\begin{abstract}
Ontology extension refers to the process of enriching an existing ontology in response to emerging requirements, making it more complete. This task is a resource-intensive and error-prone process. Large Language Models (LLMs) have shown promising performance on generating ontologies from scratch, but current approaches rarely tie ontology extension explicitly to requirements or reusable core models, and offer limited, systematic evaluation of LLM outputs.
This paper introduces \textit{OntoExtend}, a requirements-driven framework for ontology extension with LLMs. It uses retrieval-augmented generation (RAG) over relevant input ontologies and requirements in the form of competency questions to propose grounded extensions. 
We evaluate OntoExtend on 39 CQs from two use cases: a public EU-project ontology, Onto-DESIDE, and an industrial ontology from Bosch. The generated fragments show few structural issues, satisfy all functional evaluation tests, and are rated by ontology engineers as requiring minor to moderate revision before integration. These results suggest that OntoExtend is useful as a drafting assistant for requirement-driven ontology extension in real world scenarios, while remaining sensitive to CQ specificity and modelling profile.


\end{abstract}
\end{frontmatter}
\markboth{July 2026\hb}{July 2026\hb}

\section{Introduction}
\label{sec:intro}
In recent years, Large Language Models (LLMs) have been increasingly integrated into various stages of the ontology engineering pipeline. Existing work includes ontology generation with LLMs \cite{lippolis2024ontogenia,lippolis2025assessing,lippolis2025ontology,saeedizade2024navigating}, where an ontology is constructed from a set of requirements, and ontology evaluation \cite{lippolis2025large,llugiqi2025experts,tsaneva2024llm}, where the goal is to determine whether a given ontology correctly models those requirements.
However, to the best of our knowledge, no prior study has examined how LLMs can extend an existing (input) ontology by reusing ontology elements while modelling additional requirements provided in the form of competency questions (CQs), i.e. natural language questions outlining and constraining the scope of an ontology. This task, which we call \textit{ontology extension}, poses several challenges. For instance, input ontologies often contain hundreds of classes and properties, exceeding the input context limitations of current LLMs. Even when an ontology is small enough to fit within the large context window of state-of-the-art LLMs, they may be misguided by irrelevant details and produce off-target or inconsistent outputs \cite{saeedizade2024navigating}.

\begin{figure}
    \centering
    \includegraphics[width=1.00\linewidth]{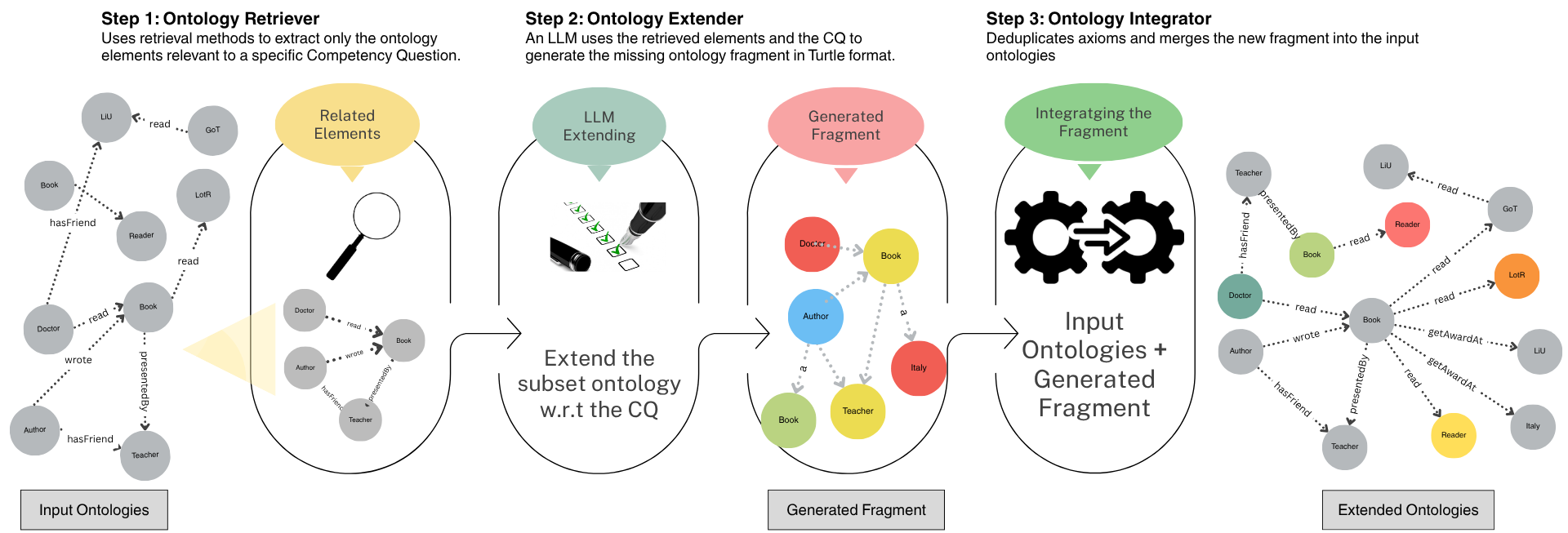}
    \caption{Overview of Ontoextend: 1) The \emph{Ontology Retriever} extracts the relevant ontology elements of the input ontologies given a new competency question. 2) The \emph{Ontology Extender} uses the retrieved ontology elements together with the competency question to prompt an LLM to generate missing ontology fragments. 3) The \emph{Ontology Integrator}  integrates the fragments into the extended ontologies.  
    }
    \label{fig:setup}
\end{figure}

Ontology extension is thus the systematic enrichment of an existing ontology to satisfy new functional requirements~\cite{pour2023phrase2onto}. Using LLMs to produce these extensions could enable incremental maintenance of ontology modules as new requirements arise. To this end, we introduce the \textit{OntoExtend} framework (Fig. \ref{fig:setup}), a retrieval-augmented approach for ontology extension that takes new CQs and the existing ontology as input. For each CQ, \textit{OntoExtend} retrieves from the input ontology the named classes and properties, along with their axioms, that are relevant to the CQ. The CQ, together with this retrieved fragment, are then provided to an LLM as relevant context without exceeding its context window or overburdening it. In the rest of this paper, we use the term \textit{fragment} to denote the self-contained set of RDF/Turtle axioms retrieved or generated for a specific extension step associated with a single CQ. This enables the LLM to generate highly contextualised ontology extensions that can be consistently integrated into the input ontology to model the new requirement.

The research questions (RQs) driving this work are: (i) Which embedding configuration is the most suitable for building a retrieval model that extracts a compact fragment of the ontology most relevant for a given CQ? (ii) Which LLMs are most effective at extending the retrieved fragment so that the CQ is correctly modelled? (iii) What evaluation criteria are required to assess the quality of the generated ontology fragments? (iv) What are the strengths and weaknesses of the extended ontologies produced by OntoExtend? To answer these questions, we propose the following contributions:
\begin{enumerate}
    \item OntoExtend: our proposed framework that extends ontologies based on new requirements\footnote{The code and the experiment data (ontologies and CQs), along with examples that could not fit in the paper, are available at \url{https://github.com/dersuchendee/OntoExtend}}; 
    \item Experiments comparing retrieval and prompting strategies to identify effective configurations for LLM-based ontology extension on two domain-specific, real-world case studies, demonstrating the approach's applicability across different domains and requirements;
    \item An evaluation methodology that assesses both structural validity and functional adequacy of generated extensions against the requirements in real-life settings through a user-based study.
\end{enumerate}

The rest of the paper is organised as follows. Section~\ref{sec:related} presents the related work; Section~\ref{sec:ontoext} describes OntoExtend; Section \ref{sec:setup} describes the experimental setup; Section~\ref{sec:evaluaiton} shows the evaluation, and Section~\ref{sec:results} reports the experimental results. Finally, Section~\ref{sec:discussion} discusses the evaluation metrics, the impact and extensibility of the benchmark, and Section~\ref{sec:conclusion} concludes the paper and outlines future research directions.

\section{Related Work}
\label{sec:related}

LLM-based ontology extension, at times called \textit{ontology enrichment}, remains in an early phase with respect to other LLM-assisted knowledge engineering tasks, with approaches ranging from interactive tools to automated enrichment systems. Early 
approaches demonstrate the potential and limitations of semi-automated methods: Soares et al. \cite{soares2025novel} extended the Agricultural Product Types Ontology (APTO) using ChatGPT-4 for interactive generation of OWL taxonomic axioms. Similarly, Matieu and Groza \cite{mateiu2023ontology} developed a Protégé plugin using a fine-tuned GPT-3 model to translate controlled natural language into OWL Functional Syntax. Work in Phrase2Onto~\cite{pour2023phrase2onto}, a prototype for ontology extension, showed promising results in this direction but was limited to toy ontologies. In addition to the mentioned limitations, these studies do not apply to larger ontologies.

Automated extension systems push toward greater autonomy while exposing new challenges: Taxoria \cite{ghamlouch2025enriching} focuses on taxonomy enrichment by having LLMs propose child terms for existing nodes, which are then validated for semantic relevance before integration with provenance tracking, though control over hallucinated nodes and implicit requirement capture remain concerns. Wu et al. \cite{wu2024ontology}'s online clustering framework extends this automation further, using LLM agents to propose and cluster new classes around existing concepts in an evolving medical ontology.
Still in the biomedical domain, Dong et al. \cite{dong2023ontology} focus on enriching OWL ontologies as formal KBs, finding that the baseline LLM-based methods are yet to achieve satisfying
performance on the benchmark.


Comprehensive extension frameworks address broader ontology engineering needs: Kholmska et al. \cite{kholmska2024enhancing}'s work extends the input ontology through a multi-LLM workflow supporting concept search, extraction, alignment, and the generation of CQs and \texttt{SPARQL} queries, though this exposes LLM limitations in highly specialised domains and requires manual repair of shallow or incorrect suggestions. Joachimiak et al. \cite{joachimiak2024artificial} developed the Artificial Intelligence Ontology using an Ontology Development Kit workflow with AI-driven curation support.
In Garcìa Fernandez et al. \cite{garcia2025ontology}, during the LLM-based extension process, two important limits were found: LLMs hallucinated when asked about existing standards and reusable ontologies, and the LLM-generated extension was shallower than the manual gold standard. Human review stayed essential at every step.

These works reveal that current LLM-based ontology extension is predominantly semi-automatic enrichment of existing inputs with a bias toward taxonomic growth, facing recurring constraints around reproducibility, reliance on interactive tools, limited support for complex axioms, and continued necessity of human judgment for requirement elicitation and validation. 
However, none of these methods focuses on the retrieval of existing ontology elements from baseline ontologies, while our approach integrates retrieval mechanisms directly into the extension process. Furthermore, most of the works use minimal to no formal evaluation.

Several related works sit nearby but are less about extending a mature input ontology and more about generating or reconstructing ontologies from scratch \cite{aggarwal2025leveraging,doumanas2025fine,huang2024large,llugiqi2025experts,plu2024comprehensive}   
A parallel line of work looks at ontology transformation and general LLM-based assistance for ontology engineering \cite{lippolis2025ontology,saeedizade2025llm_ontology_assist,saeedizade2024navigating,svatek2024welcome}. These works reinforce the usefulness of LLM support throughout the ontology lifecycle, but they do not yet address retrieval-aware, requirement-driven extension of a mature input ontology.


\begin{table*}[t]
\centering
\caption{Comparison of the related works discussed in Section~\ref{sec:related}. 
Columns indicate whether a work explicitly reports ontology pitfall analysis (OOPS!), syntax or well-formedness checking (Syntax), semantic or logical consistency validation (Consist.), verification against requirements (Req. verif.), explicit presence of superfluous elements (Superfl. el), formal user evaluation (User eval.), expert-based assessment (Expert), evidence across multiple domains (Domain gen.), evaluation on an actively used ontology (Real-world Onto.), and evidence of scalability (Scal.). 
Symbols: Y = yes, P = partial, N = not reported.}
\label{tab:comparison}

\scriptsize
\setlength{\tabcolsep}{2pt}

\resizebox{0.9\textwidth}{!}{%
\begin{tabular}{c p{2.5cm} ccc cc c cccc p{1.4cm}}
\toprule
\textbf{Ref.} &
\textbf{Approach} &
\multicolumn{3}{c}{\textbf{Structural}} &
\multicolumn{2}{c}{\textbf{Functional}} &
\makecell{\textbf{User}\\\textbf{eval.}} &
\textbf{Expert} &
\makecell{\textbf{Domain}\\\textbf{gen.}} &
\makecell{\textbf{Real-world}\\\textbf{Onto.}} &
\textbf{Scal.}  \\
\cmidrule(lr){3-5}
\cmidrule(lr){6-7}
&
&
\makecell{OOPS!} &
\makecell{Syntax} &
\makecell{Consist.} &
\makecell{Req.\\verif.} &
\makecell{Superfl. el.} &
&
&
&
&
&
\\
\midrule

\cite{soares2025novel} &
Interactive taxonomy extension
& \no & \no & \no
& \no & \no
& \yes
& \yes & \yes & \yes & \no\\

\cite{mateiu2023ontology} &
Prot\'eg\'e plugin / CNL to OWL
& \no & \no & \partialm
& \no & \no
& \no
& \no & \no & \no & \no
 \\

\cite{pour2023phrase2onto} &
Prototype ontology extension
& \no & \no & \no
& \no & \no
& \yes
& \yes & \yes & \yes & \no
 \\

\cite{ghamlouch2025enriching} &
Taxonomy enrichment (Taxoria)
& \no & \no & \no
& \no & \no
& \no
& \no & \no & \yes & \partialm
 \\

\cite{wu2024ontology} &
Online clustering framework
& \no & \no & \no
& \no & \no
& \no
& \no & \no & \yes & \yes
 \\

\cite{dong2023ontology} &
Biomedical enrichment benchmark
& \no & \no & \no
& \no & \no
& \no
& \no & \no & \yes & \partialm
 \\

\cite{kholmska2024enhancing} &
Multi-LLM extension workflow
& \no & \partialm & \partialm
& \no & \no
& \no
& \no & \no & \yes & \partialm
 \\

\cite{joachimiak2024artificial} &
AI Ontology curation support
& \no & \partialm & \yes
& \no & \no
& \no
& \no & \no & \yes & \partialm
 \\

\cite{garcia2025ontology} &
Human-reviewed ontology extension
& \no & \yes & \no
& \yes & \no
& \no
& \no & \yes & \yes & \no
 \\

\cite{huang2024large} &
RAG ontology construction
& \no & \no & \no
& \no & \no
& \no
& \no & \no & \no & \yes
 \\

\cite{aggarwal2025leveraging} &
Research ontology construction
& \no & \no & \no
& \no & \no
& \no
& \no & \no & \yes & \yes
 \\

\cite{llugiqi2025experts} &
Ontology generation
& \no & \no & \no
& \no & \no
& \no
& \no & \no & \yes & \no
\\

\cite{plu2024comprehensive} &
Ontology-Toolkit
& \no & \no & \no
& \no & \no
& \no
& \no & \no & \no & \no
 \\

\cite{doumanas2025fine} &
Ontology generation from seeds/corpora
& \no & \no & \yes
& \no & \no
& \no
& \no & \no & \no & \no
 \\

\cite{svatek2024welcome} &
Ontology transformation support
& \no & \no & \no
& \yes & \no
& \no
& \yes & \no & \yes & \no
 \\

\cite{saeedizade2024navigating} &
Ontology engineering assistant
& \no & \yes & \no
& \yes & \no
& \no
& \no & \yes & \no & \no
 \\

\cite{lippolis2025ontology} &
Ontology engineering assistant
& \yes & \yes & \no
& \yes & \yes
& \yes
& \no & \yes & \yes & \no
 \\


\rowcolor{gray!12}
Ours &
OntoExtend
& \yes & \yes & \yes
& \yes & \yes
& \yes
& \yes & \yes & \yes & \yes \\

\bottomrule
\end{tabular}%
}
\end{table*}

\section{The OntoExtend framework}
\label{sec:ontoext}
The proposed system implements a retrieval-based pipeline for ontology extension, organised into three principal subsystems: the Ontology Retriever, the Ontology Extender, and the Ontology Integrator, illustrated in Fig.~\ref{fig:setup}. Throughout this paper, input ontologies and reference ontologies refer to the same artefacts: the ontology (or ontologies) the user wants to extend, which are also indexed by the Ontology Retriever for element retrieval. The RAG component indexes the same ontology files that constitute the extension target, so that generated fragments are grounded in and consistent with the existing modelling choices. Optionally, previously generated fragments can be re-indexed so that later CQs build on earlier ones.


\subsection{Ontology Retriever}

Since the whole set of input ontologies may not fit into the context size of current LLMs, the Ontology Retriever is responsible for constructing and querying a semantic index over the input ontologies (i.e., the ontologies to be extended).

At preprocessing time, the retriever component parses each ontology file specified as input by the user and iterates over all declared OWL entities, including classes, object properties, data properties, annotations, and their SHACL shapes if available. For each entity. It constructs an \emph{OntologyElement} record containing: the entity IRI; human-readable labels and comments if present; domain and range declarations (where applicable); super- and sub-class (or sub-property) relations; and a verbatim Turtle snippet capturing the canonical declaration of the entity.

\begin{figure}
    \centering
    \includegraphics[width=1\linewidth]{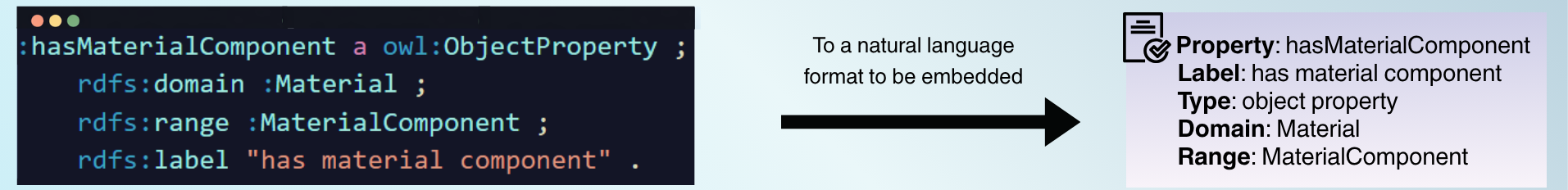}
    \caption{Example of the element to be embedded.}
    \label{fig:RAG_Elements}
\end{figure}

These OntologyElements form the input knowledge base used for subsequent retrieval.
To support semantic search, each OntologyElement is embedded into a dense vector representation using a configurable sentence embedding model. Each element is serialised as a pipe-delimited string combining its URI local name, human-readable label, comment, element type, and domain/range before embedding, e.g., `hasMaterialComponent | has material component | ... | Type: object property | Domain: Material | Range: MaterialComponent'. The resulting vectors are normalized and stored in a FAISS index \cite{douze2024faiss} configured for inner-product similarity. This choice enables efficient nearest-neighbour search over large ontology collections while preserving cosine-like similarity semantics. An example of the element to be embedded with the pipe format is shown in Figure~\ref{fig:RAG_Elements}.

At query time, a user-supplied CQ is embedded into the same vector space and used to interrogate the FAISS index. The retriever returns the top-k (the default is 20) most similar OntologyElement instances, effectively selecting those ontology entities whose textual and structural descriptions are most relevant to the CQ. This retrieval step ensures that the subsequent generation is tightly anchored in the terminology and modelling patterns of the reference ontologies. Moreover, it is the key for promoting reuse of existing ontology elements.


\subsection{Ontology Extender}


The Ontology Extender assembles the final prompt for the LLM with the task to propose consistent extensions. Retrieved elements are grouped by their source ontology and rendered as Turtle snippets, which are injected into the prompt as read-only context. In addition, it constructs a unified prefix block that merges the namespace declarations of all input ontologies, ensuring that every element IRI retrieved from any source ontology resolves correctly in the generated output and can be directly reused by the LLM without namespace conflicts. OntoExtend also enables ontology engineers to configure different prompts so that users can select among the most suitable templates at run-time.




Before a fragment is accepted, it is passed through a two-stage validator: a Turtle parser verifies syntactic correctness, and a constraint checker verifies whether generated properties satisfy the modelling conventions required in the selected use case. In the two profiles used in this evaluation, this includes explicit `rdfs:domain' and `rdfs:range' declarations, but this is a configurable governance constraint and not a universal criterion for ontology correctness. Fragments failing validation are either retried or flagged.

Given the assembled prompt, this Ontology Extender calls the LLM service and extracts the returned Turtle code block, which is treated as a candidate TBox fragment. 

\noindent\textbf{Prompt variants and deployment profiles:}
To support heterogeneous deployment requirements, the system maintains different prompt templates that configure the extender stage.
These prompts are cleanly decoupled from the input retrieval, generation, and integration pipeline. 

In our work, we used two distinct templates according to the domain of the data we experimented with (see Section 4). For the industrial ontology, we develop a SHACL-based prompt template, whereas for the EU ontology we defined a prompt template for the generic use of restrictions and reuse of classes policy.
The former instructs the LLM to output SHACL NodeShape and PropertyShape definitions following specific naming conventions and including \texttt{sh:name} labels; this is essential for downstream compliance workflows that consume SHACL artefacts. Its modelling guidance largely stops after requiring appropriate domain and range declarations. In contrast, the latter prompt template requires that every newly introduced class and property be annotated with both \texttt{rdfs:label} and \texttt{rdfs:comment}. Both impose strong reuse constraints, instructing the LLM to reuse existing ontology elements without redeclaring it. By externalising these prompt configurations, the system can switch between companies and use cases without any changes to the underlying system components.
\vspace{0em}

\subsection{Ontology Integrator}

The Ontology Integrator module is responsible for incorporating the generated fragment into the input ontologies. Its operation is limited to deduplication (removing repeated axioms, named classes, properties, and prefixes) followed by concatenation of the cleaned fragment with the input ontologies.

The Ontology Integrator additionally forwards each generated fragment to the Ontology Retriever for indexing, thereby reducing the likelihood that the Ontology Extender would recreate similar elements (classes or properties) when subsequent CQs were processed. Re-indexing allows each generated fragment to serve as a reference for subsequent CQs, promoting cross-fragment consistency. However, this was disabled in our evaluation so that each CQ fragment could be assessed independently.

\section{Experimental setup}
\label{sec:setup}
In this section, we describe the experimental setup, including the dataset we created, parameter tuning for the retrieval and the extender module, and the evaluation criteria.



\noindent\textbf{Dataset Creation:}
The evaluation covers two settings: four modules of the Onto-DESIDE ontology network\footnote{\url{https://cordis.europa.eu/project/id/101058682}}, treated jointly as one setting (20 CQs), and an internal Bosch ontology (19 CQs). Onto-DESIDE addresses circular economy data interoperability, whereas the Bosch setting follows a more manufacturing industrial profile. The dataset was manually created by systematically removing selected classes and their related properties from each ontology, and then formulating CQs about the removed elements. The procedure is outlined below:

\begin{enumerate}
    \item Select a random set of classes \(C\) from the ontology.
    \item For each class \(c\), remove \(c\) and add to the set \(P\) all properties whose domain or range restrictions references \(c\).
    \item Iteratively expand the removal process by adding \(c\) all subclasses of each removed class to \(C\), and include in \(P\) any properties that reference these subclasses. This step is repeated iteratively until no additional subclasses are found.
    \item Finally, construct a set of CQs associated with \(C\) and \(P\). Each CQ is formulated by asking what each corresponding class with its properties, were intended to represent or accomplish within the original ontology.
\end{enumerate}

\noindent
This procedure yields a set of CQs that systematically probe the missing ontology components, enabling an effective evaluation of the OntoExtend framework.

\noindent\textbf{Dataset statistics}: The EU-project parts have 20 CQs in total, while the Industry ontology comprises 19 CQs. Detailed counts of tokens, the sum of classes and properties, and axiom counts for each subset are provided in the GitHub repository. The input ontologies were converted to Turtle with compact prefix declarations to reduce size. Token counts are measured using the GPT-4o tokeniser (\url{https:// platform.openai.com/tokenizer}), see Table \ref{tab:token-sizes}.

\begin{table}[h]
\centering
\caption{Token counts of input ontologies after conversion to Turtle.}

\begin{tabular}{lccc}
\toprule
Ontology fragment & Size (tokens) & Axioms & Classes+Properties \\
\midrule
EU-project — part 1 & 75\,000& 2920& 405 \\
EU-project — part 2 & 22\,000& 958&242  \\
EU-project — part 3 & 25\,000&1172 &270  \\
EU-project — part 4 & 6\,000&228 &54  \\
Industry use case & 22\,000& 979&  134\\

\bottomrule
\end{tabular}
\label{tab:token-sizes}
\end{table}


\noindent\textbf{Parameter and configuration tuning}

We ran the experiment on a separate subset to find out the best embedding model and parameter configuration to test on the main experiment, along with the best prompt to minimise the common mistakes by LLMs in ontology generation.

\noindent\textbf{Best retrieval configuration} 

To embed each ontology element into a vector with good settings, we first ran a small experiment to choose the embedding model and text configuration before the main study to find a good set of parameters and embedders. Basing on the results of previous research \cite{lippolis2025assessing,saeedizade2024navigating}, we compared three OpenAI embedding models
: \texttt{text-embedding-3-small}, \texttt{text-embedding-3-large}, and \texttt{text-embedding-ada-002} under four different styles of formulating ontology elements as a raw text before sending them into embedding mode. In particular, we varied the delimiter used to separate consecutive elements, comparing the use of a pipe character (|) with the use of a newline character (\verb|\n|). For each configuration, we manually judged relevance on five CQs not present in the test data and computed precision at 3 and 20 (focusing on top-3 retrieved for answer quality in case of creating a small ontology subset, but still monitoring top-20 to have an estimate of recall). In particular, we computed precision at cutoffs 3 and 20 for the retrieved elements being tagged as similar to the input CQ by the Ontology Retriever. Evaluations are done manually by two ontology engineers cross cross-checking each other's work, letting $P_3$ and $P_{20}$ denote precision at 3 and 20 (often referred to as P@3 and P@20). Our primary selection metric is a weighted average $M_{\mathrm{w}} = 0.7\,P_3 + 0.3\,P_{20}$
which emphasises the quality of the very top-ranked results (those most likely to be consumed by the LLM in Ontology Extender) while still rewarding configurations that retrieve more relevant items in the top 20. As an additional check, we also monitored the multiplicative average of them $M_{\mathrm{prod}} = P_3 \cdot P_{20}$
,
which penalises configurations that perform well at only one cutoff. Both metrics agreed on the same best configuration per model.

Across all 12 configurations, the best overall result came from \texttt{text-embedding- ada-002} with pipe-separated element axioms and comments included. On this basis, we selected that as the embedding model for the main experiment. Inputs of the top configurations are shown in Table~\ref{tab:embedding-comparison}.

\begin{table}[h]
\centering
\caption{Embedding model comparison results by examining whether to include comments in the embedding and how to separate axioms in the raw text. }

\resizebox{0.85\textwidth}{!}{
\begin{tabular}{l@{\hspace{2.5em}} l@{\hspace{2.5em}}c@{\hspace{1.5em}} c}
\hline
Embedding model        & Separator, Comments         
& $M_{\mathrm{prod}}$ & $M_{\mathrm{w}}$ \\
\hline
\texttt{text-embedding-3-small} & Newline
, with Comments    
& 0.22                & 0.62             \\
\texttt{text-embedding-3-large} &Newline
, without Comments 
& 0.20                & 0.54             \\
\texttt{text-embedding-ada-002} & Pipe, with Comments       
& \textbf{0.23}                & \textbf{0.63 }            \\
\hline
\end{tabular}
}
\label{tab:embedding-comparison}
\end{table}


\noindent\textbf{Best prompt and LLM extender}

For the LLMs, we selected \texttt{o1-preview}, reported as the best-performing model in previous work~\cite{lippolis2025ontology}, and \texttt{GPT-5}, the latest LLM from OpenAI. We did not select other models or families because previous studies reported lower performance relative to \texttt{o1-preview} for what concerns ontology generation tasks~\cite{lippolis2025ontology}.

Prompt development followed the same tuning procedure used for retrieval. After fixing the retrieval configuration and integrating it into the extension framework, we evaluated several prompt variants on a small development set of CQs (disjoint from the main evaluation set) to identify formulations that minimised typical LLM errors in ontology construction. The initial prompt was modelled closely on the prompt used in previous work~\cite{lippolis2025ontology} and was augmented with an explicit list of observed pitfalls not to be included in the output. This baseline was refined iteratively: after each pilot run, two ontology engineers inspected generated axioms for recurrent problems and adjusted the prompt wording, constraints accordingly.

The final prompt comprises (i) a general template that specifies the task and required output format, (ii) a section into which the Ontology Retriever’s elements are inserted, (iii) a placeholder for the given CQ, and (iv) a directive that indicates the expected modelling formalism. The framework populates all components automatically, while the style directive may be supplied by the user to constrain output style. For instance, in the EU-Project, this section requested OWL restrictions for creating restrictions in the generated ontologies; in the industry use case, we disallowed OWL restrictions and required only SHACL shapes. Consequently, the two use cases should not be interpreted as testing the same level of OWL expressivity. The EU-project setting evaluates OWL-style fragment generation, whereas the industry setting evaluates extension under a SHACL-oriented engineering profile used in that deployment context.
The final prompt template is available on the Github repository.

\subsection{Evaluation criteria}
\label{subsec:eval_criteria}
Building on previous studies in LLM-assisted knowledge engineering~\cite{lippolis2025ontology}, we adopt a multidimensional evaluation setup that combines structural and functional metrics along with human evaluation. 

\textbf{Structural Metrics}. We employ the Ontology Pitfall Scanner (OOPS!) \cite{poveda2014oops} for reporting on pitfalls, Pellet reasoner and syntax checking through the RDFLib Python library\footnote{\url{https://rdflib.readthedocs.io/en/stable/}}.

\textbf{Functional Metrics.} To assess the generated ontologies, we employ CQ verification \cite{blomqvist2012ontology} and counting of superfluous elements based on Lippolis et al. \cite{lippolis2025ontology} by modifying the definition of the latter for the specific task of ontology extension. CQ verification evaluates whether a given CQ is actually represented in the ontology by attempting to formulate a \texttt{SPARQL} query that retrieves an answer for that CQ. If no such \texttt{SPARQL} query can be written to obtain an answer, the CQ is considered \emph{not modelled} in the ontology.
Likewise, \cite{lippolis2025ontology} defines a superfluous element in a way that treats some generated classes and properties 
as superfluous because they cannot be referenced directly in the \texttt{SPARQL} query used for CQ verification. In our work, we adopt a more specific definition: a \emph{superfluous element} is a named class or property that: 1) is not mentioned in the verification \texttt{SPARQL} used to in CQ verification, and 2) is not connected to any component appearing in that query by a \texttt{
  subClassOf} or \texttt{
  subPropertyOf} relation.
Under this definition, elements that are absent from the verification query but are hierarchical relatives (subclasses or subproperties) of query components are \emph{not} considered superfluous, while genuinely disconnected named classes and object properties are flagged as superfluous.

\label{subsec:survey}
 \textbf{Survey.} Finally, we created a survey to ask six ontology engineers from both academia and industry to evaluate the extended ontologies. We adopted evaluation criteria adapted from Monka et al.~\cite{monka2025enhancing}, where two dimensions for evaluating \textit{SPARQL} generation are proposed: \emph{Correctness} and \emph{Completeness}. 
In our work, \emph{Correctness} measures the syntactic and semantic quality of the generated fragment in isolation, and \emph{Completeness} measures the sufficiency of the generated fragment after it has been combined with the input ontology; it captures the amount of effort required by an ontology engineer to make the fragment usable.
Users rated the generated ontology fragments on a five-point Likert scale along two dimensions, namely \emph{Correctness} and \emph{Completeness}.

For \emph{Correctness}, a rating of 1 indicates a generation failure, for example because the output exceeded token limits. A rating of 2 denotes a syntactically erroneous fragment, such as one containing issues in the generated code. A rating of 3 is assigned when the fragment is syntactically correct but semantically incorrect, for instance due to an inappropriate taxonomy. A rating of 4 corresponds to a fragment that is both syntactically and semantically correct, but does not fully satisfy the intent of the competency question (CQ). Finally, a rating of 5 indicates a fragment that is syntactically and semantically correct and fully aligned with the intended meaning of the CQ.

For \emph{Completeness}, a rating of 1 means that the generated ontology fragment is not useful and would require complete manual reworking. A rating of 2 indicates that significant changes are needed, corresponding roughly to 25--40\% of the fragment requiring modification. A rating of 3 reflects the need for moderate changes, with approximately 11--25\% of the fragment needing revision. A rating of 4 denotes that only minor changes are required, typically affecting around 1--10\% of the fragment. A rating of 5 indicates that the generated fragment, in combination with the input ontology, is complete and correct.



\section{Evaluation}
\label{sec:evaluaiton}
In this section, we present the methodology used to evaluate the generated ontology extensions based on the criteria introduced in Section~\ref{sec:setup} in detail.

\begin{figure}
    \centering
    \includegraphics[width=1\linewidth]{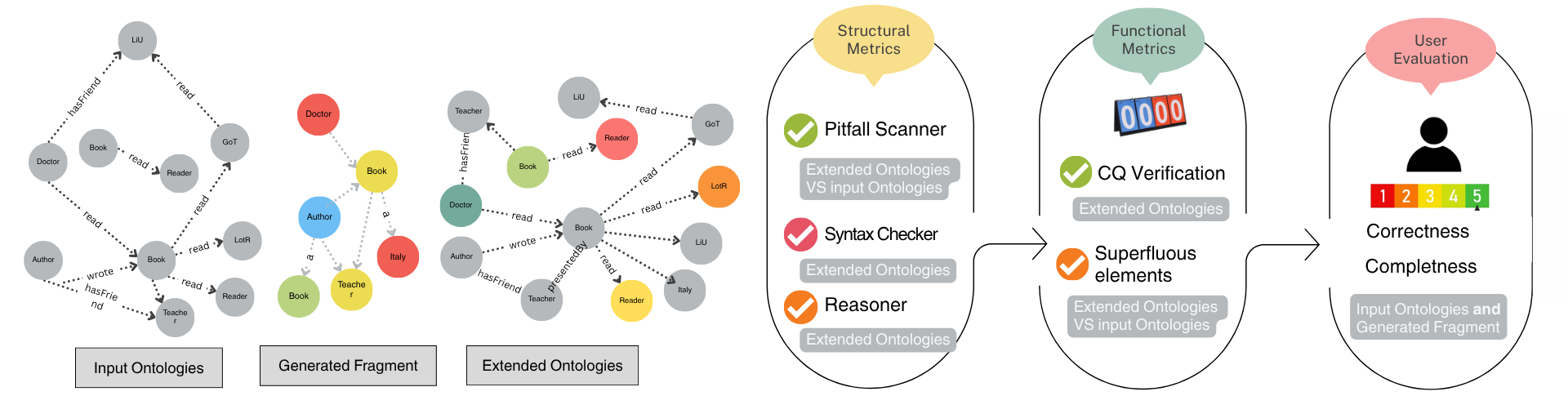}
    \caption{Evaluation workflow: input ontologies, the generated fragments, and the extended ontologies are fed to different evaluation metrics to create the final report. }
    \label{fig:eval}
\end{figure}
\subsection{Structural Evaluation}

As shown on the right side of Fig.~\ref{fig:eval}, each generated extension module is validated for correct Turtle syntax via the RDFLib Python library. To assess common modelling pitfalls, we run OOPS! on the input ontology both before and after integrating the generated extension, compare the reported pitfalls, and document any new issues introduced by the extension. 
In addition, we execute the Pellet reasoner to perform a consistency check on the input ontology after the generated extension has been integrated.

\subsection{Functional Evaluation}

\paragraph{\textbf{Functional adequacy (CQ verification)}.} As described in Section \ref{subsec:eval_criteria}, we perform CQ verification on each generated extension after it is added to the input ontology to ensure that each CQ is correctly modelled (see Figure~\ref{fig:eval}). Two ontology engineers independently verify the CQ modelling decisions and then cross-check each other's annotations; any disagreement is resolved through discussion. This adjudication process follows the same general workflow as in the work by \cite{lippolis2025assessing}.

\paragraph{\textbf{Superfluous elements}.} After CQ verification, the same pair of engineers inspects the integrated file to identify and count superfluous elements introduced by the generated extension by comparing the number of superfluous elements before and after adding the extension. As detailed in Section \ref{subsec:eval_criteria}, each decision is cross-checked, and we report the number of superfluous elements that were generated but deemed unnecessary by the OntoExtend framework.

\subsection{Evaluation by Ontology Engineers}

In the experiment, we recruited six ontology engineers from both industry and academia to evaluate the generated extension modules. After a short briefing that explained the task and the survey form
, each evaluator received a set of CQs, the input ontologies, and the generated ontology fragments for each CQ. For every generated extension, the evaluators answered the two survey questions regarding correctness and completeness (see section \ref{subsec:survey}). The evaluators were also encouraged to provide free-text comments. Finally, the evaluators also participated in a debriefing session to share general feedback. Participants were free to use any tools they preferred to visualise the ontology artefacts (e.g., Protégé, Topbraid EDG, VSCode).

After data collection, we computed the mean correctness and completeness metrics from the surveys. We also computed the \emph{Fleiss} weighted observed agreement $P_o$ \cite{moons2025_fleiss} to quantify the inter-annotator agreement among our evaluators.


\section{Results}
\label{sec:results}

\vspace{-0.5em}
In this section, we present the results of our experimental setup according to the evaluation criteria outlined in Section \ref{sec:evaluaiton} of the paper.

\subsection{Results of Structural Evaluation}
Firstly, the generated ontology fragments almost did not exhibit any syntax issues in Turtle.
Second, the analysis of the integrated ontologies with OOPS! shows that the generated extension fragments do not introduce any new \emph{critical} or \emph{important} modelling pitfalls after being integrated into their input; only a small number of minor issues were observed in specific cases. For the EU-project ontology, we found two types of minor issues. First, P02 (creating synonyms as classes) appears in a single EU-project only: in that case, each generated extension contributed one P02 instance. Second, P04 (unconnected ontology elements) was detected in just two of the four EU-project use cases; in those two cases, each extension added at most one additional P04 instance.

For the industry ontology, OOPS! reported P08 (missing annotations). This behaviour is likely due to the annotation-free style of the input ontology: OntoExtend appears to have replicated that style as it is asked in the prompt, yielding missing annotation warnings when annotations were expected by OOPS!. On average, P08 occurs approximately 3.7 times per CQ-based extension for both LLMs.

Comparing OntoExtend to existing work on ontology generation~\cite{fathallah2024neon,lippolis2024ontogenia,lippolis2025ontology}, which showed a considerable number of \textit{important} and \textit{critical} OOPS! pitfalls, OntoExtend exhibits 
only a few minor pitfalls. Overall, the results indicate that the generated extensions are structurally acceptable and do not systematically introduce new modelling defects; the limited minor issues observed are confined to specific use cases and are straightforward to fix.
\vspace{-0.5em}

\subsection{Results of Functional Evaluation}

\begin{table}[t]
\centering
\caption{Summary of structural (syntax and OOPS!) and functional (CQ-verification and percentage of superfluous elements) evaluation for the EU-Project and the Industry ontologies, all LLMs combined and separated.}
\resizebox{\textwidth}{!}{
\begin{tabular}{lcccc}
\toprule
Use case & \begin{tabular}[c]{@{}c@{}}OOPS! P2,P4\&P8\\  (o1-prev,GPT-5)\end{tabular} & \begin{tabular}[c]{@{}c@{}}CQ-verification\\ (o1-prev,GPT-5)\end{tabular} & \begin{tabular}[c]{@{}c@{}}Syntax errors\\ (o1-prev,GPT-5)\end{tabular} & \begin{tabular}[c]{@{}c@{}}Superfluous elements\\ (o1,GPT-5)\end{tabular} \\

\midrule
EU-Project    & 13 (7,6) 
& 100\% (100\%, 100\%) & 0\% (0\%,0\%) & 2\%  (3.8\%,0\%) \\
Industry & 141(70,71) 
& 100\% (100\%,100\%) & 2.5\% (5\%,0\%) & 0\% (0\%,0\%)  \\
\bottomrule
\end{tabular}}
\label{tab:ce-bosch-syntax}

\end{table}

The results of CQ verification show that all CQs in both use cases are correctly modelled, with no minor issues of the type identified by Saeedizade and Blomqvist~\cite{saeedizade2024navigating}. Both \texttt{GPT-5} and \texttt{o1-preview} produce extension fragments with only negligible amounts of superfluous elements.
A comparison of the number of superfluous elements with the work in~\cite{lippolis2025ontology} further highlights this improvement. Using our refined definition of superfluous components, the ontology fragments generated in this work contain fewer than 2\% unnecessary elements, whereas~\cite{lippolis2025ontology} reported around 30\% superfluous elements in their generated ontologies (35\% with~\cite{lippolis2025ontology}'s original definition of superfluous elements).
\vspace{-0.5em}

\subsection{Results of survey from Engineers' Evaluation}

Six ontology engineers, comprising three engineers for the EU-Project and three who worked on the industry ontology, evaluated the generated ontology fragments by filling out the survey, results shown in Table~\ref{tab:combined-iaa}.


\begin{table}[t]
\centering
\caption{Survey results on evaluating ontology extension fragments for the EU-Project and industry ontologies. The ratings are between 1 and 5 (see section \ref{subsec:eval_criteria})}
\resizebox{0.75\textwidth}{!}{

\begin{tabular}{l@{\hspace{2em}} l@{\hspace{2em}} l@{\hspace{1.5em}} c@{\hspace{1.5em}} c@{\hspace{1.5em}} c}
\toprule
Model & Metric &
\multicolumn{2}{c}{Industry} &
\multicolumn{2}{c}{EU-Project} \\
\cmidrule(lr){3-4} \cmidrule(lr){5-6}
& & Mean  & \emph{Fleiss} $P_o$ & Mean & \emph{Fleiss} $P_o$ \\
\midrule
\texttt{o1-preview} & Correctness   & 4.91 & 0.97 & 3.69 & 0.80 \\
\texttt{o1-preview} & Completeness  & 4.56 & 0.89 & 3.11 & 0.85 \\
\texttt{GPT-5}      & Correctness   & 4.96 & 0.98 &3.66  & 0.87 \\
\texttt{GPT-5}      & Completeness  & 4.54 & 0.87 &  2.94& 0.87 \\
\bottomrule
\end{tabular}}
\label{tab:combined-iaa}
\end{table}

Engineers who evaluated fragments of the EU ontologies identified some systematic deficiencies. First, they observed unconnected elements within fragments that ought to have been linked to the input ontologies, e.g. via a \texttt{subClassOf} relation. Second, class naming was often suboptimal (for example, excessively specific names), indicating recurrent poor modelling patterns.
Model-specific issues were also reported. Outputs produced by \texttt{GPT-5} frequently omitted explicit domain and range declarations for newly introduced properties, imposed incorrect semantic restrictions (e.g., inappropriate use of \texttt{allValuesFrom} or \texttt{complementOf}), and employed overly simplified names. Conversely, fragments generated by \texttt{o1-preview} commonly redefined existing classes, contained syntactically incorrect axioms, which were not shown by the tools used for the Structural evaluation, assigned incorrect domains to some properties, and created object properties that appeared too specific compared to individual CQs. Overall, the evaluators with high agreement concluded that the generated fragments require moderate revision before they can be safely integrated into the input ontologies.

Regarding the industry ontology,  the main errors concerned that in few cases, only \texttt{SHACL} property shapes were added without the corresponding object property plus proper \texttt{rdfs:domain} and \texttt{rdfs:range} definitions, and in a few cases the \texttt{sh:datatype} was missing. Furthermore, it was noted that sometimes LLMs omitted the required \texttt{rdfs:subClassOf} axioms and lacked comment annotations for classes and properties. This is valid for both models. Specific only to \texttt{o1-preview}, it tends to introduce many additional Property shapes and object properties that are not needed to answer the CQs. There is also concern in two comments that extensions were generated without grounding from the CQ or the input ontologies. In this case, the LLM anticipated some extensions purely based on patterns available in the input ontology. Overall, however, the results regarding the industry ontology show a high degree of user satisfaction (minor or no changes needed to the fragments) and strong observed agreement among the evaluators.


\vspace{-0.5em}

\section{Discussion}
\label{sec:discussion}
\vspace{-0em}

In this section, we analyse the results, outline the main limitations of our study and experimental setup, and discuss directions for future work.


\noindent\textbf{Overall results:} Overall, our results show that OntoExtend can reliably produce high-quality ontology extensions across both domains we have considered. The generated fragments are almost always syntactically valid, introduce only a small number of minor OOPS! pitfalls, especially when compared with existing related work. They correctly model all evaluated CQs, and contain fewer than 2\% superfluous components, while the six ontology engineers generally judged them to be syntactically and semantically adequate and requiring at most minor to moderate edits. Taken together, these findings indicate that OntoExtend already offers a practically useful level of automation for real-world ontology extension and maintenance. In practice, OntoExtend functions as a drafting assistant: it generates a complete extension module that users typically only need to review and lightly edit. This keeps human–tool interaction minimal, shifting effort from manual axiom authoring to quick validation and targeted refinement. This significantly lowers the barrier to entry for domain experts, allowing them to expand ontologies with less reliance on often lacking ontology engineers.

\noindent\textbf{Formulation and quality of the CQs affecting the results:}
Industry ontology engineers assessed the industry ontology and the EU-project ontologies (the latter developed by academic partners), and inspected the generated extension for the EU-project. Respondents were encouraged to add free-text notes on each generated module.

Our evaluation covered two distinct usage scenarios: (i) extending an ontology from highly specific, pre-determined CQs (the tool acts as an assistant for writing axioms), and (ii) constructing or extending ontologies from more general, open CQs (the tool must make substantive modelling choices).
Although the same overall procedure was used to create the datasets, the EU-project CQs differ qualitatively from the industry CQs. The latter were deliberately very focused and targeted: they were authored jointly by domain experts and ontology engineers so that high-level modelling choices were already fixed and each CQ primarily required the tool to generate a missing class or a small set of properties from the natural-language question. 
By contrast, the CQs for the EU-Project ontology were defined in a more open manner. I.e., the authors, mostly domain experts, left many of the modelling decisions to the tool. Consequently, the tool in this case has to make the open modelling decisions when generating the ontology fragments.     

This difference in CQ style affects the evaluation. For the EU-project CQs, the LLMs often produced reasonably structured extensions, but ontology engineers judged some outputs as less complete because additional manual refinement was needed to align the results with modelling choices implied by the full specification. Thus, lower correctness and completeness scores in this setting reflect rather the limitations of the CQ definition (i.e., missing clarity on the expectations and contextual detail), not necessarily a weakness of the OntoExtend framework. In the industry setting, the tool mainly needed to render valid axioms according to an input ontology and hence achieved much higher scores for completeness and correctness.
The lower scores on correctness and completeness observed in the EU-project scenario from the evaluators' survey results can therefore be explained as a result of the different task complexity of the ontology generations. 

\noindent\textbf{LLM behaviour as an implicit quality measure:} 
As discussed previously, the same framework is applied to two qualitatively different CQ sets and associated ontologies: one in which CQs are tightly scoped, and another in which CQs lack clear expectations and context, leaving many modelling decisions open. 
Our analysis shows that OntoExtend tends to perform better in terms of correctness and completeness, alignment with expert expectations, and reduced need for post-editing when the CQs and background ontologies are well structured, precise, and internally coherent. 
These findings in turn suggest that LLM performance can serve as a proxy indicator for the quality of requirements (CQs) and/or ontological artefacts: if a given set of CQs and input ontologies systematically yields higher quality extensions with fewer modelling errors, then this combination is in a pragmatic sense ``better'' for downstream ontology engineering with LLMs (and probably also humans). Conversely, when the same solution struggles, this most likely is a result of ambiguity, under-specification, or hidden modelling assumptions in the input. In line with recent discussions on using LLMs as auxiliary evaluators in knowledge engineering, our results therefore support the interpretation of LLM behaviour as a novel, tool-centric dimension of ontology and requirement quality.

\noindent\textbf{Practical impact:} Sending very large ontology fragments (e.g., $\sim$75k-token) directly to an LLM resulted in substantially longer end-to-end response times and higher API costs in our experiments. By contrast, OntoExtend retrieves a compact subset of the ontology and provides only that subset to the LLM; this greatly reduces the typical LLM turnaround and lowers the amount of data transmitted and the associated cost. 

\noindent\textbf{Limitations and future work:} Despite the promising aspects of this work, it also has limitations. First, our evaluation is restricted to two domain-specific use cases; a broader set of domains and modelling scenarios would be necessary to draw more general conclusions. Second, we experimented with only a small number of LLM configurations, and a more systematic comparison across different models and providers could reveal important performance differences. A further potential threat to the validity of our results is data leakage, i.e., the possibility that the underlying LLMs have been exposed during pre-training to ontologies, CQs, or related documentation that are similar to (or overlap with) our experimental material. In our setup, for the EU-Project ontology, the ontologies have been publicly released, but the material concerning CQs and their coverage is not publicly available. For the industry ontology, it was not disseminated in public code or data repositories, which reduces the risk that related artefacts have been seen during model training. Future work will explore more rigorous safeguards against leakage, such as on-the-fly benchmark generation, experiments with open-weight models trained on controlled corpora, and dedicated checks for overlap between evaluation data and known pre-training sources.
Finally, our findings indicate that the quality and formulation of requirements (in our case, CQs) have a substantial impact on the outcome. Future work should explicitly analyse and improve CQ quality itself, for example by (semi-)automatically generating CQs, or mapping them to existing CQ templates (see the work by Keet et al. \cite{keet2019claro}).

\section{Conclusion}
\label{sec:conclusion}
In this work, we introduced OntoExtend, a scalable requirement-driven framework that extends one or more input ontologies according to the requirements specified in a set of 
CQs. OntoExtend processes each CQ individually and extracts, for each CQ, the relevant existing ontology elements as context for an LLM to generate the missing ontological structures necessary to fully answer the CQ in a coherent manner. 
To answer our research questions, we found that using \texttt{text-embedding-ada-002} as an embedder produced the most effective retrieval subsets for our experiments (RQ1). Furthermore, both \texttt{GPT-5} and \texttt{o1-preview} yielded comparable performance when supplied with the retrieved context (RQ2): the generated fragments contained almost no Turtle syntax errors, did not introduce any new \emph{critical} or \emph{important} OOPS! pitfalls, only a negligible amount of minor pitfalls, and exhibited only a small fraction of superfluous elements. We propose a multi-faceted evaluation (RQ3), combining syntactic checks, OOPS! analysis, CQ verification, measurement of unnecessary classes/properties (superfluous), and an analysis of the generated outputs by six ontology engineers. This evaluation setup shows that OntoExtend is able to extend ontologies that are both correct and complete according to ontology engineers. 
Overall, for what concerns the benefits and weaknesses of the extended ontologies produced by OntoExtend (RQ4), the extended fragments are structurally and syntactically correct, accurately model all evaluated CQs, and add fewer than 2\% superfluous elements, with engineers often needing only minor edits in the industry use case. The positive evaluation from ontology engineers shows high potential for the usage of OntoExtend in real scenarios. However, we still observe weaknesses such as occasional missing domain/range declarations, suboptimal or overly specific names, and some unconnected elements. Moreover, performance degrades when CQs are open-ended or underspecified, showing that OntoExtend is effective when requirements are precise but sensitive to the quality of the input CQs.

\footnotesize{
\noindent\textbf{Acknowledgements.} This project has received funding from the European Union’s Horizon Europe research and innovation programme under grant agreements no. 101058682 (Onto-DESIDE) and the Swedish Vinnova-funded project SwePass (Dnr. 2024-02504). 
This work was also supported by the PhD scholarship ``Discovery, Formalisation and Re-use of Knowledge Patterns and Graphs for the Science of Science'', funded by CNR-ISTC through the WHOW project (EU CEF programme - grant agreement no. INEA/CEF/ICT/ A2019/2063229) and COST Action CA23147 GOBLIN – Global Network on Large-Scale, Cross-domain and Multilingual Open Knowledge Graphs, supported by COST (European Cooperation in Science and Technology, https://www.cost.eu). 


\noindent\textbf{Use of Generative AI.} ChatGPT was used to enhance the readability of some of the text and improve the language of this paper, after the content was first added manually. After using this service, the authors reviewed and edited the content as needed and take full responsibility for the content of the published article.

\smallbreak

\noindent\textbf{Disclosure of Interests.}
The authors have no competing interests to declare that are
relevant to the content of this article.}

\bibliographystyle{splncs04}
\bibliography{bibliography}




\end{document}